\documentclass[10pt,twocolumn,letterpaper]{article}

\usepackage[pagenumbers]{iccv} 
\usepackage{multirow}
\usepackage{graphicx}  
\usepackage{multirow}  
\usepackage{booktabs}  
\usepackage{amssymb}

\definecolor{iccvblue}{rgb}{0.21,0.49,0.74}
\usepackage[pagebackref,breaklinks,colorlinks,allcolors=iccvblue]{hyperref}

\def\paperID{3} 
\def\confName{ICCV}
\def\confYear{2025}

\title{Long-Tailed Data Classification by Increasing and Decreasing Neurons During Training
}

\author{Taigo Sakai\\
Meijo-University\\
1-501 Shiogamaguchi, Tempaku-ku, Nagoya 468-8502, Japan\\
{\tt\small 200442066@ccalumni.meijo-u.ac.jp}
\and
Kazuhiro Hotta\\
Meijo-University\\
1-501 Shiogamaguchi, Tempaku-ku, Nagoya 468-8502, Japan\\
{\tt\small kazuhotta@meijo-u.ac.jp
}
}

\begin{document}
\maketitle
\begin{abstract}
\vspace{-2em}

In conventional deep learning, the number of neurons typically remains fixed during training. However, insights from biology suggest that the human hippocampus undergoes continuous neuron generation and pruning of neurons over the course of learning, implying that a flexible allocation of capacity can contribute to enhance performance. Real-world datasets often exhibit class imbalance—situations where certain classes have far fewer samples than others, leading to significantly reduce recognition accuracy for minority classes when relying on fixed size networks.
To address the challenge, we propose a method that periodically adds and removes neurons during training, thereby boosting representational power for minority classes. By retaining critical features learned from majority classes while selectively increasing neurons for underrepresented classes, our approach dynamically adjusts capacity during training. Importantly, while the number of neurons changes throughout training, the final network size and structure remain unchanged, ensuring efficiency and compatibility with deployment.
Furthermore, by experiments on three different datasets and five representative models, we demonstrate that the proposed method outperforms fixed size networks and shows even greater accuracy when combined with other imbalance-handling techniques. Our results underscore the effectiveness of dynamic, biologically inspired network designs in improving performance on class-imbalanced data.
\end{abstract}
\section{Introduction}
\label{sec:intro}

The balance of data, in which the number of samples in each class is approximately equal, has a significant impact on the performance of machine learning models.
Generally, when trained on well-balanced data, where each class has a similar number of samples, the model can learn all class characteristics equally, making it easier to achieve high classification accuracy.

In balanced datasets, each class contributes equally, allowing the model to learn uniformly and reduce bias. This leads to more consistent classification performance, even if class difficulty varies.

\begin{figure}[t]
\centering
\includegraphics[width=1.0\linewidth]{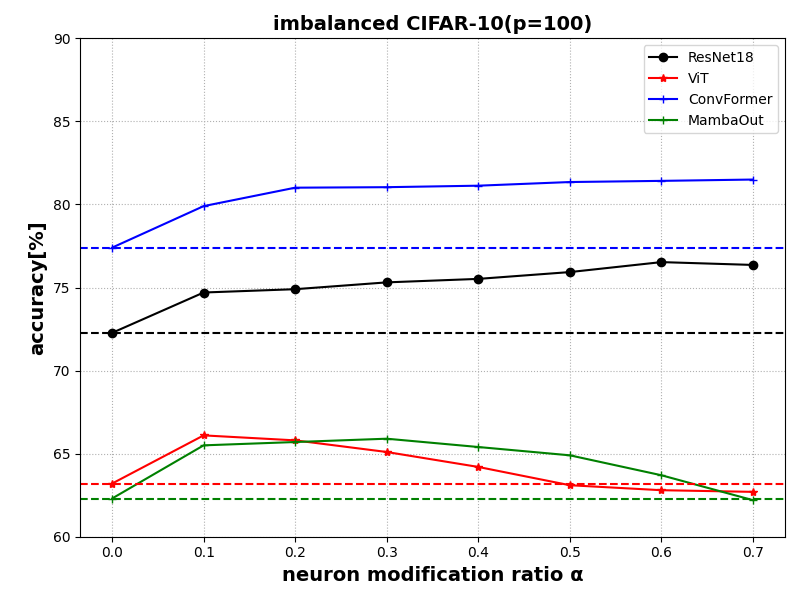}
\caption{Overall accuracy on the Imbalanced CIFAR-10 dataset (imbalance factor $p=100$) for different architectures. The horizontal axis represents the neuron modification ratio $\alpha$, which determines the fraction of neurons dynamically added or removed during training. The vertical axis shows classification accuracy (\%). The dashed lines indicate baseline accuracies without neuron modification. We show the results for ResNet18 (black), ViT (red), ConvFormer (blue), and MambaOut (green). While ResNet18 and ConvFormer show a slight increase in accuracy as $\alpha$ increases, ViT and MambaOut experience performance degradation, suggesting that the impact of neuron modification varies depending on the architecture.
}
\label{fig:neuron_modification}
\end{figure}

However, in many real-world applications, datasets are often imbalanced, meaning that the number of samples in certain classes is dramatically smaller compared to others.
For instance, in the medical domain, it is common to encounter only a small number of samples for a particular disease, while a large number of healthy samples are readily available.
Moreover, in anomaly detection or fraud detection tasks, an enormous amount of normal data is typically collected, whereas only a tiny fraction corresponds to anomalies or fraudulent cases.
Such biased distributions reflect inherent properties of natural and societal phenomena.


To mitigate the imbalance, various approaches have been proposed, such as resampling (\textit{oversampling} and \textit{undersampling}) and cost-sensitive weighting which often referred to as \textit{Cost-Sensitive Learning}.
In oversampling methods like \textit{SMOTE}~\cite{chawla2002smote} and \textit{ADASYN}~\cite{he2008adasyn}, additional synthetic samples are generated for the minority class, whereas undersampling techniques remove part of the majority class to achieve a more balanced dataset.
Furthermore, representative examples of specialized loss functions are \textit{Focal Loss}~\cite{lin2017focal}, \textit{logit adjustment Loss}~\cite{menon2021longtaillearninglogitadjustment}, and \textit{Adaptive t-vMF Dice Loss}~\cite{kato2023adaptive}, and so on. These loss functions assigns larger weights to samples that are harder to classify, thus promoting focused learning on difficult or minority samples.
Although these methods have achieved certain levels of success, they also pose risks.
Oversampling can produce synthetic samples that deviate from the true distribution, while undersampling may discard valuable information about the majority class.

From the perspective of human learning mechanisms, it is widely suggested that the brain flexibly rewires its neural circuits by pruning synaptic connections~\cite{lin2020dynamic, li2016pruning, article, gadhikar2023random, unconstrained-channel-pruning} and growing new neurons as necessary~\cite{wu2021firefly, wang2023learning, evci2022gradmax, yao2024masked}.
In contrast, conventional deep learning models typically fix their network structure including the number of layers and neurons, without adapting to the demands of imbalanced data.
Although some recent researches~\cite{gurbuz2022nispa, gurbuz2023sharpsparsityhiddenactivation, gurbuz2024nice} 
in continual learning have explored the methods to dynamically expand or shrink the network by adding or removing neurons, their direct application to class imbalance remains limited.
Therefore, in this study, we propose dynamically adjusting the network structure itself by increasing or decreasing neurons to tackle imbalanced data from an alternative viewpoint, complementing conventional resampling and weighting strategies.

In this paper, we introduce a method to grow necessary neurons and remove unnecessary ones based on the magnitude of the loss gradient.
Specifically, if certain neurons prove highly influential on the errors associated with minority classes, we add new neurons to strengthen representation learning for those classes.
Conversely, neurons that appear to offer negligible contribution are pruned to reduce overall redundancy, thus trimming excess parameters.
Our goal is to create a learning environment in which even severely imbalanced data can effectively improve the performance on minority classes.

Our method can be applied not only to VGG~\cite{7486599} and ResNet~\cite{he2016deep} but also to a variety of other architectures such as Vision Transformer (ViT)~\cite{dosovitskiy2021an}, MetaFormer~\cite{yu2022metaformeractuallyneedvision}, and MambaOut~\cite{yu2024mambaout}, thereby enhancing its versatility.


Through experiments conducted on three benchmark datasets (Imbalanced CIFAR-10/100~\cite{Cui2019ClassBalancedLB}, ImageNet-LT~\cite{liu2019large}, and iNaturalist-2018~\cite{vanhorn2018inaturalist}) and models of five different architecture types (VGG, ResNet, ViT, MetaFormer, MambaOut), we demonstrate that the proposed method outperforms fixed-size networks and further improves accuracy.
Figure ~\ref{fig:neuron_modification} presents the impact of the neuron modification ratio on classification accuracy for different architectures trained on the Imbalanced CIFAR-10 dataset ($p=100$). 
Figure shows the potential of dynamically adjusting neurons to enhance model performance.

Our primary contributions are summarized as follows.
\begin{enumerate}
    \item We propose a new method for imbalance learning that dynamically adjust the number of neurons during training.
    \item Our approach enhances the representation of minority classes while preserving critical features of majority classes, thereby mitigating the effects of class imbalance.
    \item We conducted experiments on three benchmark datasets, and five models, showing that the proposed method outperforms fixed-size networks and further improves accuracy.
\end{enumerate}

The structure of this paper is as follows. First, Section ~\ref{sec:Related} provides the overview of conventional methods for addressing class imbalance and discusses related studies on dynamically modifying network structures during training, positioning our research within this context. Section ~\ref{sec:Method} details the proposed method, which involves dynamically increasing and decreasing the number of neurons during training, and explains how the network structure adapts accordingly. Section ~\ref{sec:Experiment} presents experimental results on three datasets and five representative models, comparing our approach with fixed-size networks and other imbalance mitigation techniques, followed by an analysis of the effectiveness of the proposed method. Finally, Section ~\ref{sec:Conc} concludes our study.
\section{Related works}
\label{sec:Related}

\subsection{Class Imbalanced Learning}

There is a wide variety of approaches for handling imbalanced data\cite{buda2018systematic, DBLP:journals/air/ChenYYSC24}.
Resampling methods (e.g. oversampling or undersampling) are widely used due to their ease of implementation and their direct adjustment of class ratios.
However, oversampling risks generating synthetic samples that may deviate from the original distribution and can increase computational costs, 
while undersampling risks discarding important majority-class samples.
By contrast, \textit{cost-sensitive learning} prioritizes minority classes by adjusting the loss calculation within the model, and several techniques have been proposed to increase the focus on challenging samples\cite{lin2017focal, cui2019class, zhang2018cost, kato2023adaptive}.

Although these methods can be effective to some extent, they are not universally applicable to every task. 
In fact, it has been reported that their performance benefits often plateau when the class distribution is extremely skewed\cite{kang2021learning}.
On the other hand, by increasing or decreasing task-specific neurons using a gradient, which is an universal criterion, our method can be applied to a variety of tasks and models.

\subsection{Previous Attempts to Modify Network Structure}

In recent years, research has advanced toward dynamically altering the scale of deep learning models during training.
For example, \textit{Firefly} (Firefly Neural Architecture Descent) expands the model in an increasing direction by conducting a localized search for optimal network growth.
However, its experiments are limited to VGG, and no applications to class-imbalanced data have been reported.
\textit{NISPA}\cite{gurbuz2022nispa} and \textit{NICE}\cite{gurbuz2024nice} are designed for \textit{continual learning} and include mechanisms for pruning unnecessary neurons and connections to accommodate new information.
Nonetheless, their verifications commonly employ ResNet, and no published results explicitly address class imbalance.
Additionally, in SInGE\cite{yvinec2022singe}, the effect of each neuron on the network output can be evaluated over the long term by integrating the gradient change over time. This method is reported to improve the accuracy of pruning by selecting more appropriate neurons without being influenced by short-term gradient fluctuations.
Although these approaches incorporate the concepts of neuron addition or removal, they do not directly target imbalanced learning, setting them apart from the focus of this research.

Therefore, in imbalanced classification, we adjust the number of neurons based on gradient magnitude to suppress overfitting. To properly account for minority-class contributions, we accumulate gradients over an epoch rather than a single batch. Additionally, we weight minority-class gradients to enhance their impact on neuron selection.

\subsection{Position of This Research}

Our aim is to extend the concept of dynamically adjusting network structures by revising its specifically to imbalanced learning.
Concretely, we estimate neuron importance based on gradient information, adding neurons deemed necessary for the minority classes while pruning those that contribute little to learning. 
This allocation of parameters specifically emphasizes error correction for underrepresenting classes, potentially capturing features that traditional resampling or cost-sensitive methods fail to detect.
Moreover, our experiments demonstrate that the proposed method can be integrated not only with VGG and ResNet but also with different architectures such as \textit{ViT}
and \textit{MambaOut}, 
underscoring its broad applicability.
A key contribution of this work is its explicit verification on imbalanced datasets, a context in which approaches like Firefly, NISPA, and NICE have not been thoroughly examined, thereby showcasing novel effectiveness through these comparative evaluations.

\section{Proposed Method}
\label{sec:Method}

In this section, we explain the details of our method for handling class-imbalanced data by dynamically adjusting the number of neurons during training. Our method comprises four main components: 
\textbf{(1)}~neuron selection and modification based on loss gradients, 
\textbf{(2)}~gradient reweighting to mitigate class-imbalance effects, 
\textbf{(3)}~weight scaling to stabilize training after neuron modification, 
\textbf{(4)}~gradient diversity assurance through random initialization of BatchNorm parameters for newly added neurons.
We also describe the gradient computation process in both CNNs and MLPs, which underpins our neuron selection strategy. 

\begin{figure}[t]
    \centering
    \includegraphics[width=0.8\linewidth]{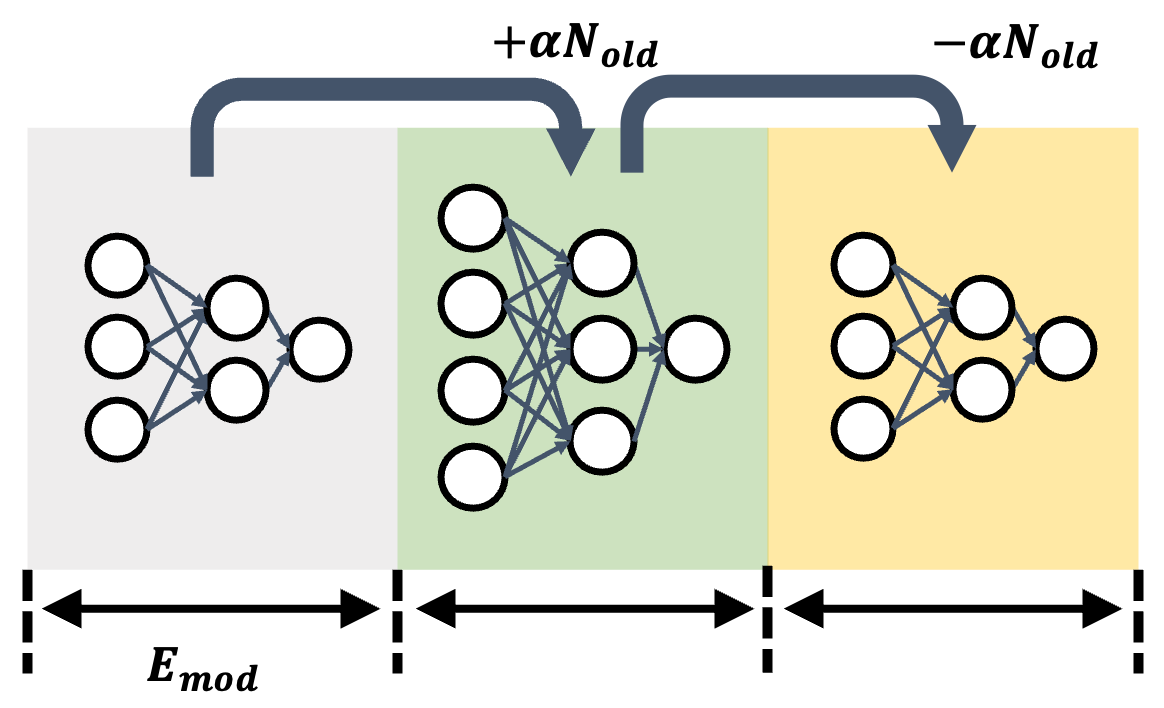}
    \caption{Conceptual illustration of our proposed method. 
      We dynamically adjust the number of neurons throughout training.
      In the first phase (gray), we start with an initial network. 
      In the second phase (green), the network is expanded or pruned based on gradient information.
      In the third phase (yellow), further adjustments are made to handle imbalanced data effectively.}
    \label{fig:concept_figure}
\end{figure}

Figure~\ref{fig:concept_figure} shows the conceptual flow of our proposed method.
We start with an initial network (gray region) where all neurons remain fixed during the early training phase.
At predetermined epochs, we move to the second phase (green), in which additional neurons are introduced or unnecessary neurons are pruned based on gradient information.
Finally, in the third phase (yellow), we refine the network structure further to handle imbalanced data effectively.
The following subsections detail each step of this process, including how we compute the gradient-based importance, how many neurons to add or remove, and how we re-initialize BatchNorm parameters.

\subsection{Neuron Selection and Modification}
\label{subsec:NSM}
\subsubsection{Motivation for Dynamic Neuron Adjustment}

Deep learning models are often trained with a fixed number of layers and neurons, an approach that may be suboptimal when dealing with highly imbalanced datasets. In such datasets, majority-class samples usually dominate the learned representations, hindering the effective learning of minority-class features. To address this, we propose dynamically modifying the network structure, increasing neurons that are crucial for minority-class learning while pruning those that appear less relevant.

\subsubsection{Timing of Neuron Modification and Gradient Accumulation}

We introduce a parameter \( E_{\text{mod}} \) to specify how frequently neuron modifications are performed. In particular, once training enters the \( E_{\text{mod}} \)-th epoch, we start to accumulate gradient information over all mini-batches. By the end of that epoch, we compute a running average of these gradients. This strategy ensures that minority-class gradients are not overlooked, 
even if certain mini-batches contain few or no minority-class samples. Once the epoch finishes, we execute our neuron addition and deletion in a paired manner (one addition per one deletion). As a result, the \emph{total} number of neurons remains nearly the same, preserving an overall balance akin to what is observed in biological neural systems.

\subsubsection{Gradient-Based Neuron Selection}

Let \(\theta\) be the set of parameters of the neural network, and let \(L\) be the loss function. For a mini-batch \(b\) containing \(n\) samples, we define a batch weight \(w_b\) (detailed in Section~\ref{subsec:reweighting}) to reflect class imbalance. The gradient of the loss with respect to the parameters associated with neuron \(i\) is 
\begin{equation}
G_i = \frac{1}{B} \sum_{b=1}^{B} \Bigl(w_b \cdot \nabla_{\theta_i} L_b\Bigr),
\end{equation}
where \(B\) is the number of mini-batches in an epoch.  
We rank all neurons by the magnitude of \(G_i\). 
Those with the highest gradient magnitudes are considered necessary (and thus added), 
while those with the smallest magnitudes are pruned.  
This approach was empirically found to be more effective than pruning neurons based on \(\ell_1\)-norm (i.e., weight magnitude). We hypothesize that gradient magnitude more directly captures the \emph{contribution} of a neuron to reduce the loss, especially for minority classes.

\subsubsection{Hyperparameter-Controlled Neuron Expansion and Reduction}

To control the scale of neuron modification, we introduce a hyperparameter \(\alpha\) (where \(0.0 \leq \alpha \leq 1.0\)) to determine the fraction of neurons to be added or removed at each modification step. Formally, if a layer has \(N_{\text{old}}\) neurons, we compute
\begin{equation}
N_{\text{new}} = N_{\text{old}} + \alpha N_{\text{old}},
\end{equation}
for the neurons to be added, and we remove an equal number based on the smallest gradient magnitudes.  
By pairing the addition and deletion counts, the total number of neurons remains effectively unchanged after each modification step, mirroring the approximate constancy of total neuron counts in the human brain.  
Experimentally, we found that \(\alpha \approx 0.3\) yields stable and robust performance across various datasets and models, although the precise optimal value may vary slightly depending on the specific task.
The reason why \(\alpha \approx 0.3\) is optimal is that if $\alpha$ is too large, the network undergoes significant structural changes, leading to unstable learning. Conversely, if $\alpha$ is too small, the adjustment becomes insufficient, resulting in limited effectiveness. Experimental results indicate that $\alpha$ around $0.3$ achieves a balanced modification of neurons, ensuring stable learning while contributing to improved accuracy.

\subsection{Gradient Reweighting for Imbalanced Data}
\label{subsec:reweighting}

\subsubsection{Class Imbalance and Loss Gradient Scaling}

Imbalanced datasets often have loss gradients dominated by majority-class samples, making it challenging to learn minority-class features. To mitigate this, we assign a per-class weight \( w_c \) as
\begin{equation}
w_c = \frac{N_{\max}}{N_c},
\end{equation}
where \(N_c\) is the number of samples in class \(c\), and \(N_{\max}\) is the sample size of the largest class. 
This boosts the gradient contribution of minority classes.

\subsubsection{Mini-Batch Level Weighting}

For a mini-batch \(b\) with \(n\) samples, we define the batch weight \( w_b \) as
\begin{equation}
w_b = \sum_{i=1}^{n} w_{c_i},
\end{equation}
where \(c_i\) is the class label of the \(i\)-th sample. We incorporate \(w_b\) when computing the aggregated gradient
\begin{equation}
G_i = \frac{1}{B} \sum_{b=1}^{B} \Bigl(w_b \cdot \nabla_{\theta_i} L_b\Bigr).
\end{equation}

\paragraph{Synergy with Neuron Expansion}

By reweighting minority-class samples, their associated gradients become more prominent. Consequently, neurons important for these samples display larger gradient magnitudes and are thus more likely to be selected for expansion. This synergy is key to ensure that the network adaptively dedicates sufficient capacity to underrepresented classes.

\subsection{Gradient Computation in CNNs and MLPs}

\subsubsection{CNN Case}

For a convolutional neural network (CNN), let \(W^{(l)}\) be the kernel weights in the \(l\)-th convolutional layer, and \(X^{(l)}\) denotes the input tensor to the layer. The forward pass is as
\begin{equation}
Y^{(l)} = W^{(l)} * X^{(l)},
\end{equation}
where \(*\) indicates the convolution operation. The gradient of the loss with respect to \(W^{(l)}\) is
\begin{equation}
\nabla_{W^{(l)}} L = X^{(l)} * \nabla_{Y^{(l)}} L.
\end{equation}
We aggregate these gradients across mini-batches, and at the end of each relevant epoch, we use the accumulated information for neuron selection.

At the end of each epoch, we rank neurons based on \(|G_i|\) and apply our neuron selection strategy: neurons with the highest gradient magnitudes are added, while those with the lowest are pruned. This approach ensures that the model dynamically adapts its capacity to focus on underrepresented classes.

\subsubsection{MLP Case}

In a multilayer perceptron (MLP), each neuron in layer \(l+1\) is fully connected to the neurons in layer \(l\). Let \(h^{(l)}\) be the activation at the layer \(l\) and \(W^{(l)}\) the weight matrix. The forward pass is as 
\begin{equation}
h^{(l+1)} = W^{(l)} \, h^{(l)}.
\end{equation}
The gradient of the loss with respect to \(W^{(l)}\) is as 
\begin{equation}
\nabla_{W^{(l)}} L = h^{(l)} \cdot \nabla_{h^{(l+1)}} L.
\end{equation}
Similar to the CNN case, we gather these gradient updates over mini-batches and then consolidate them to guide our neuron addition and removal steps.

\subsection{Weight Scaling for Stability}
\label{subsec:WS}
\subsubsection{Scaling Strategy}

Once we decide to modify the number of channels (in CNNs) or neurons (in MLPs), we apply a scaling factor $s$ to stabilize the magnitude of the weights. Specifically, when the channel count changes from \(C_{\text{old}}\) to \(C_{\text{new}}\), we define
\begin{equation}
s = \sqrt{\frac{C_{\text{old}}}{C_{\text{new}}}},
\end{equation}
and update the weight matrix as
\begin{equation}
W' = s \cdot W.
\end{equation}
This scaling avoids abrupt shifts in parameter distributions after neuron modification, contributing to smoother training convergence.

\subsection{Ensuring Gradient Diversity with BatchNorm Randomization}
\label{subsec:GDA}

\subsubsection{BatchNorm Parameter Initialization}

For newly added neurons, we randomize the learnable BatchNorm parameters \(\gamma\) and \(\beta\) around He-initialization\cite{7410480} to ensure a diverse set of initial states. Specifically,
\begin{equation}
\gamma' = \gamma + \epsilon, \quad \beta' = \beta + \epsilon,
\end{equation}
where \(\epsilon \sim \mathcal{N}(0,\sigma^2)\) is a small perturbation. This strategy prevents all newly created neurons from starting with identical statistics, promoting better gradient diversity and reducing the risk of poor local minima.

\subsection{Extension to Transformer-Based Architectures}

Although our primary illustrations focus on CNNs and MLPs, we also apply the same approach to Transformer-based models such as ViT and MambaOut. In these architectures, we \emph{do not} modify the number of attention heads. Instead, we change the hidden dimension (equivalent to channels in CNNs) within each head. By treating these dimensions in a manner analogous to CNN channels, we can consistently select and prune neurons (or channels) based on the same gradient-based criterion. This design choice retains the overall structure of multi-head attention while allowing the model to focus resources on underrepresented classes.

\subsection{Summary of the Proposed Method}

Our method dynamically modifies the network structure to handle class imbalance, integrating the following steps.
\begin{enumerate}
    \item \textbf{Neuron Selection and Modification}:
        Neurons are periodically evaluated at epoch boundaries based on accumulated gradients. We add neurons with large gradient magnitudes and remove those with the smallest magnitudes.
    \item \textbf{Gradient Reweighting}:
        We assign per-class weights to balance the loss contributions from minority classes. This reweighting amplifies minority-class gradients, supporting neuron expansion where needed.
    \item \textbf{Gradient Computation}:
        We handle both CNNs and MLPs by aggregating layer-wise gradients, then apply the same principle to Transformer-based models by focusing on hidden dimensions rather than head counts.
    \item \textbf{Weight Scaling}:
        Weight matrices are rescaled to stabilize training after channel or neuron count changes, ensuring minimal disruption to the learned representations.
    \item \textbf{BatchNorm Randomization}:
        Newly created neurons receive BatchNorm parameters initialized with small random perturbations around He-initialization, enhancing gradient diversity and reducing the risk of converging to poor local minima.
\end{enumerate}

These components interactively act to boost minority-class performance by targeting critical neurons, reallocating network capacity, and maintaining training stability even under severe class imbalance.


\section{Experiment}
\label{sec:Experiment}

We conducted experiments to verify the effectiveness of the proposed method on the imbalanced CIFAR-10, CIFAR-100, ImageNet-LT, and iNaturalist-2018 datasets.

\subsection{Experimental Setup}

For CIFAR-10 and CIFAR-100, we introduced artificial class imbalance by setting the imbalance factor $p$ to 10 and 100. 
We used VGG19, ResNet18, ResNet50, ConvFormer, CAFormer, ViT, and MambaOut as backbone networks. These architecture were trained by using CELoss and AdamW with a learning rate of 0.1 and momentum of 0.9. In the conventional method, training was conducted with a fixed number of neurons. 
In our experimental results in Table~\ref{tab:cifar-results}, ~\ref{tab:imagenet-lt-results}, ~\ref{tab:inaturalist2018-results}, ~\ref{tab:neuron-selection-results}, "Conventional" refers to the standard training approach where the network structure remains fixed throughout training, without any dynamic neuron modifications.

In contrast, our proposed method dynamically increased or decreased neurons based on the importance estimated by the exponential moving average of gradients.
For CIFAR-10 and CIFAR-100, neurons were adjusted every 30 epochs, modifying up to $\alpha \cdot 100$\% of neurons over 150 epochs. For ImageNet-LT and iNaturalist-2018, adjustments were made every 20 epochs, modifying up to $\alpha \cdot 100$\% of neurons over 100 epochs. Model performance was evaluated based on the average test accuracy across all classes.

\subsection{Results}
\textbf{Results on Imbalanced CIFAR-10/100}
Table~\ref{tab:cifar-results} presents the classification accuracy on the imbalanced CIFAR-10 and CIFAR-100 datasets under two levels of class imbalance factor $p$. 
Across all network architectures and both datasets, our proposed method consistently outperforms conventional training. Our method demonstrates significant performance improvements, particularly in highly imbalanced settings. For instance, in CIFAR-100, ResNet18 achieves an increase from 37.9\% to 41.8\%, and ConvFormer-s18 improves from 56.2\% to 60.1\%, both showing a 3.9 percentage gain. Similarly, in CIFAR-10, ResNet18 sees an increase from 72.2\% to 75.4\%, while CAFormer-s18 improves from 77.4\% to 81.5\%. These results suggest that our method effectively enhances the learning process by dynamically adjusting neurons to better represent minority-class features.
Accuracy for all methods improves when the imbalance ratio is less severe ($p=10$) compared to the more severe setting ($p=100$). However, the consistent performance gains of our method in both cases indicate that neuron modification is beneficial regardless of the degree of class imbalance. Furthermore, CNN-based models, such as ResNet18 and ConvFormer-s18, tend to benefit the most from our approach, while transformer-based models like ViT and MambaOut also show accuracy improvements.

\begin{table}[t]
    \centering
    \begin{minipage}[t]{0.5\textwidth}
        \caption{Comparison of accuracy(\%) on imbalanced CIFAR-10/100.}
        \label{tab:cifar-results}
        \centering
        \scalebox{0.72}{
            \begin{tabular}{c|c|c|c|c|c}
            \hline
            \multirow{2}{*}{\textbf{Dataset}} & \multirow{2}{*}{\textbf{Network}} & \multicolumn{2}{c|}{\textbf{$p = 0.01$}} & \multicolumn{2}{c}{\textbf{$p = 0.1$}} \\
            \cline{3-6}
             &  & {conventional} & {Ours} & {conventional} & {Ours} \\ \hline
            \multirow{3}{*}{CIFAR-10}
             & VGG19 & 70.6 & \textbf{72.9} & 83.4 & \textbf{84.5} \\
             & ResNet18 & 72.2 & \textbf{75.4} & 84.3 & \textbf{85.6} \\
             & ConvFormer-s18 & 82.4 & \textbf{84.9} & 84.8 & \textbf{85.8} \\
             & CAFormer-s18 & 77.4 & \textbf{81.5} & 80.4  & \textbf{81.0} \\
             & ViT-B & 63.2 & \textbf{65.1} & 74. & \textbf{76.3} \\
             & MambaOut-femto & 62.3 & \textbf{65.9} & 74.1 & \textbf{75.6} \\ \hline
             
            \multirow{3}{*}{CIFAR-100}
             & VGG19 & 44.3 & \textbf{46.2} & 62.7 & \textbf{64.1} \\
             & ResNet18 & 37.9 & \textbf{41.8} & 63.8 & \textbf{65.1} \\
             & ConvFormer-s18 & 56.2 & \textbf{60.1} & 63.2 & \textbf{64.2} \\
             & CAFormer-s18 & 55.6 & \textbf{59.5} & 61.7 & \textbf{62.3} \\
             & ViT-B & 35.8 & \textbf{39.1} & 42.3 & \textbf{44.5} \\
             & MambaOut-femto & 31.5 & \textbf{33.9} & 34.3 & \textbf{36.0} \\ \hline
            \end{tabular}
        }
    \end{minipage}
    \vspace{1.5em}
    \vfill
    \centering
    \begin{minipage}[t]{0.5\textwidth}
        \caption{Comparison of accuracy(\%) on ImageNet-LT.}
        \label{tab:imagenet-lt-results}
        \centering
        \scalebox{0.8}{
        \begin{tabular}{c|c|c|c|c|c}
            \hline
            \textbf{Method} & \textbf{Network} & \textbf{Overall} & \textbf{Many} & \textbf{Medium} & \textbf{Few} \\ 
            \hline
            \multirow{5}{*}{Conventional} 
            & ResNet50 & 40.3 & 48.1 & 39.8 & 19.1 \\ 
            & ConvFormer-s18 & 45.7 & 52.5 & 41.5 & 22.3 \\ 
            & CAFormer-s18 & 43.8 & 50.7 & 40.3 & 19.8 \\ 
            & ViT-B & 37.3 & \textbf{56.2} & 30.5 & 10.8 \\ 
            & MambaOut-femto & 33.1 & 50.4 & 27.2 & 8.8 \\ 
            \hline
            
            \multirow{3}{*}{Ours} 
            & ResNet50 & \textbf{41.9} & \textbf{49.0} & \textbf{40.8} & \textbf{22.1} \\ %
            & ConvFormer-s18 & \textbf{48.2} & \textbf{52.5} & \textbf{44.7} & \textbf{26.1} \\
            & CAFormer-s18 & \textbf{45.2} & \textbf{51.2} & \textbf{42.9} & \textbf{12.7} \\
            & ViT-B & \textbf{41.1} & 56.1 & \textbf{35.7} & \textbf{17.6} \\ %
            & MambaOut-femto & \textbf{35.7} & \textbf{52.5} & \textbf{30.2} & \textbf{10.6} \\ \hline %
            \end{tabular}
        }
    \end{minipage}
    \vspace{1.5em}
    \vfill
    \centering
    \begin{minipage}[t]{0.5\textwidth}
        \caption{Comparison of accuracy(\%) on iNaturalist-2018.}
        \label{tab:inaturalist2018-results}
        \centering
        \scalebox{0.8}{
            \begin{tabular}{c|c|c|c|c|c}
            \hline
            \textbf{Method} & \textbf{Network} & \textbf{Overall} & \textbf{Many} & \textbf{Medium} & \textbf{Few} \\ 
            \hline
            \multirow{3}{*}{Conventional} 
            & ResNet50 & 62.2 & 61.3 & 63.6 & 59.7 \\ %
            & ViT & 54.8 & 64.6 & 53.7 & 52.5 \\ %
            & MambaOut-femto & 50.6 & 60.2 & 48.9 & 47.8 \\ \hline %
            \multirow{3}{*}{Ours} 
            & ResNet50 & \textbf{63.6} & \textbf{61.6} & \textbf{64.1} & \textbf{62.1} \\ %
            & ViT & \textbf{56.8} & \textbf{65.1} & \textbf{54.7} & \textbf{55.1} \\ %
            & MambaOut-femto & \textbf{52.7} & \textbf{61.0} & \textbf{52.9} & \textbf{51.2} \\ \hline %
            \end{tabular}
        }
    \end{minipage}
\end{table}

\textbf{Results on Large-Scale Datasets: ImageNet-LT and iNaturalist-2018}

To evaluate the scalability of our method, we conducted experiments on two large-scale, long-tailed datasets: ImageNet-LT and iNaturalist-2018. The results, shown in Tables~\ref{tab:imagenet-lt-results} and~\ref{tab:inaturalist2018-results}, demonstrate that our method consistently improves classification accuracy across different architectures and class distributions.

Following \cite{menon2021longtaillearninglogitadjustment}, we record accuracy for different class groups by categorizing them into three levels: "Many", which includes classes with at least 100 training samples; "Medium", which consists of classes with between 20 and 99 training samples; and "Few", which includes classes with fewer than 20 training samples.

On ImageNet-LT, our approach improves overall accuracy across all models, with ResNet50 increasing from 40.3\% to 41.9\%, ConvFormer-s18 from 45.7\% to 48.2\%, and ViT from 37.3\% to 41.1\%. Importantly, our method significantly enhances the recognition of minority classes category: "Few",
where ResNet50's accuracy in the "Few" category improves from 19.1\% to 22.1\%, and ConvFormer-s18 increases from 22.3\% to 26.1\%. ViT improves from 10.8\% to 14.6\% in the accuracy of "Few" category. These results confirm that our method enhances feature learning for underrepresented classes by dynamically adjusting neuron capacity.

Similarly, on iNaturalist-2018, our method consistently improves classification performance. ResNet50 achieves an accuracy gain from 62.2\% to 63.6\%, ViT from 54.8\% to 56.8\%, and MambaOut-femto from 50.6\% to 52.7\%. The improvements are particularly pronounced for minority-class recognition, with ResNet50 increasing from 59.7\% to 62.1\%, ViT from 52.5\% to 55.1\%, and MambaOut-femto from 47.8\% to 51.2\%. These findings highlight that our method is effective in addressing extreme class imbalance in real-world large-scale datasets, where certain categories have very few samples.

\textbf{Comparison of Neuron Selection Criteria}

Table~\ref{tab:neuron-selection-results} compares different neuron selection criteria in terms of classification accuracy. Our method, which selects neurons based on accumulated gradients over epochs, 
achieves the highest accuracy across all datasets. The L1-based selection performs better than random selection but does not achieve the same level of improvement as gradient-based methods. Notably, selecting neurons based on final batch gradients shows improvement but remains inferior to our gradient accumulation approach, with over epochs. These results confirm that considering accumulated gradients provides a more robust criterion for neuron selection, leading to better model adaptation.

\textbf{Impact on Class-Wise Accuracy}

\begin{figure}[t]
\centering
\subfloat[Conventional]{
   \includegraphics[width=0.5\linewidth]{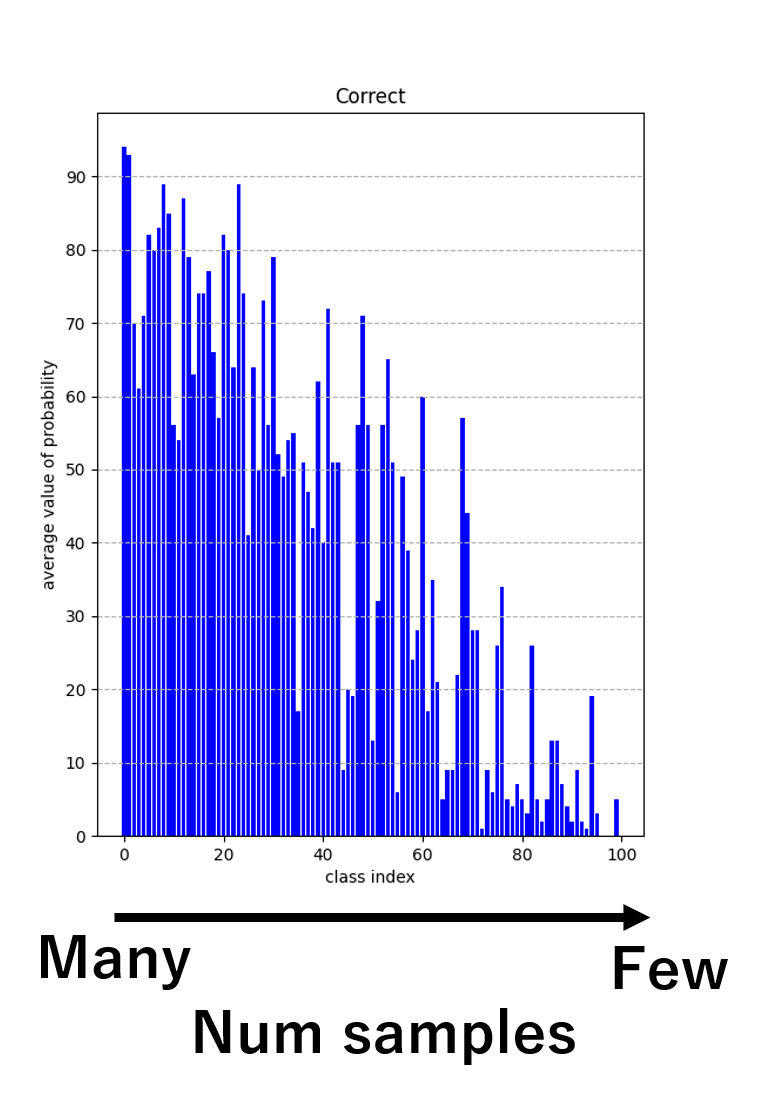}
   \label{fig:normalresnet18}
}
\subfloat[Ours]{
   \includegraphics[width=0.5\linewidth]{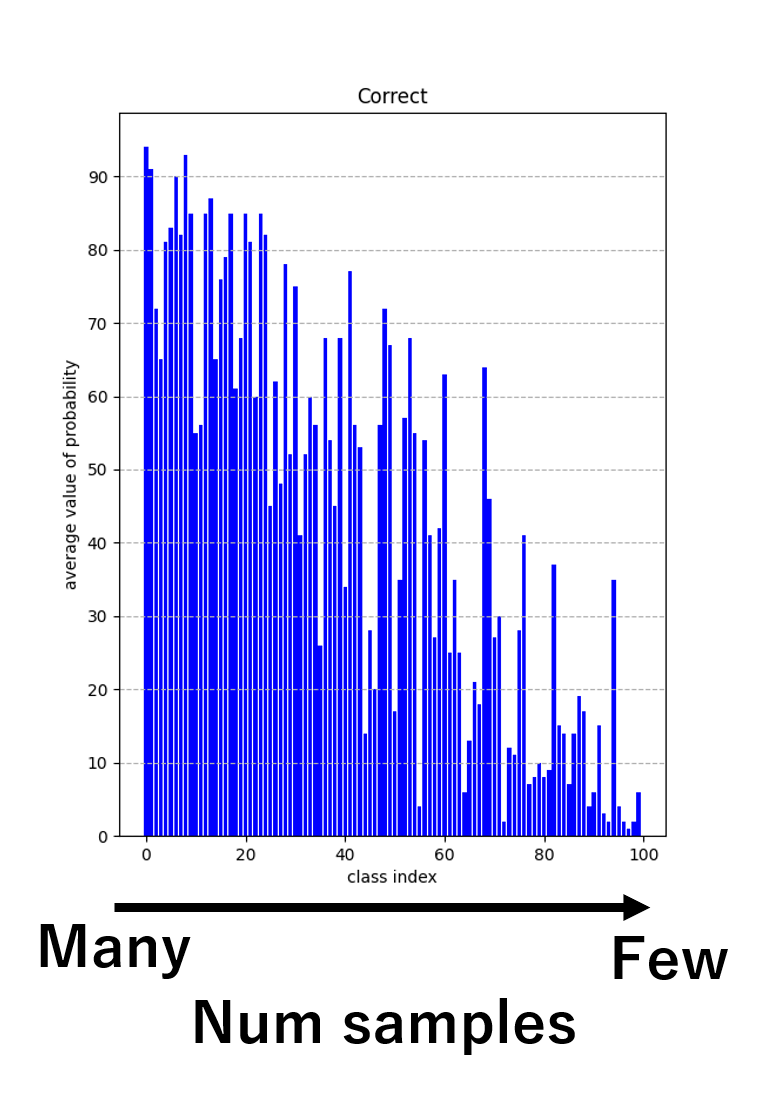}
   \label{fig:zougenresnet18}
}

\caption{The conventional training method (\ref{fig:normalresnet18}) struggles with minority classes, leading to a significant drop in accuracy for classes with fewer samples. In contrast, our method (\ref{fig:zougenresnet18}) demonstrates a more balanced accuracy distribution, particularly improving performance for underrepresented classes. 
The horizontal axis represents class indices sorted by the number of samples, from the most to the least frequent classes, while the vertical axis shows the average classification accuracy for each class. This confirms that our dynamic neuron adjustment strategy effectively mitigates the challenges of class imbalance, ensuring that minority classes are better learned without compromising majority-class accuracy.
\label{fig:classwise-accuracy}
}
\end{figure}

Figure~\ref{fig:classwise-accuracy} illustrates the impact of our method on class-wise accuracy distribution for CIFAR-100 with $p=100$ (ResNet18) . 
The conventional training method struggles with minority classes, leading to a significant drop in accuracy for classes with fewer samples. In contrast, our method demonstrates a more balanced accuracy distribution, particularly improving performance for underrepresented classes. This confirms that our dynamic neuron adjustment strategy effectively mitigates the challenges of class imbalance, ensuring that minority classes are better learned without compromising majority-class accuracy.

\begin{table}[t]
    \centering
    \caption{
        Comparison of average accuracy using different neuron selection criteria, with $\alpha=0.3$, ResNet18 on CIFAR10/100 ($p=100$), ResNet50 on ImageNet-LT.
    }
    \label{tab:neuron-selection-results}
    \scalebox{0.8}{
        \renewcommand{\arraystretch}{1.2}
        \begin{tabular}{c|c|c|c|c|c}
        \hline
        \textbf{Dataset} & {Conventional} & {Random} & {L1} & {Gradient} & {Ours} \\  & & & & {(Finaly)} & {} \\ \hline
        CIFAR-10 & 72.2 & 72.1 & 73.5 & 73.9 & \textbf{75.4} \\
        CIFAR-100 & 37.9 & 37.8 & 39.7 & 39.2 & \textbf{41.8} \\
        ImageNet-LT & 37.3 & 37.2 & 37.2 & 37.6 & \textbf{40.9} \\ \hline
        \end{tabular}
    }
\end{table}

\textbf{Ablation study}

\begin{table}[t]
    \centering
    \caption{Abation study of average accuracy, with $\alpha=0.3$, ResNet18 on CIFAR10/100 ($p=100$).}
    \begin{tabular}{c c c c | c c}
        \hline
        NAM & GR & WS & GDA & CIFAR-10 & CIFAR-100 \\
        \hline
        $\times$ & $\times$ & $\times$ & $\times$ & 72.2 & 37.9 \\
        $\checkmark$ & $\times$ & $\times$ & $\times$ & 74.3 & 39.9 \\
        $\checkmark$ & $\checkmark$ & $\times$ & $\times$ & 74.5 & 40.7 \\
        $\checkmark$ & $\times$ & $\checkmark$ & $\times$ & 73.9 & 40.3 \\
        $\checkmark$ & $\checkmark$ & $\checkmark$ & $\times$ & 74.9 & 41.0 \\
        $\checkmark$ & $\checkmark$ & $\checkmark$ & $\checkmark$ & \textbf{75.4} & \textbf{41.8} \\
        \hline
    \end{tabular}
    \label{tab:results}
\end{table}

Table~\ref{tab:results} shows an ablation study using ResNet18 on CIFAR-10/100 ($p=100$), evaluating the contribution of each component in our method: Neuron Selection and Modification (NAM) in section \ref{subsec:NSM}, Gradient Reweighting (GR) in section \ref{subsec:reweighting}, Weight Scaling (WS) in section \ref{subsec:WS}, and Gradient Diversity Assurance (GDA) in section \ref{subsec:GDA}.

Starting from the baseline (72.2\% on CIFAR-10, 37.9\% on CIFAR-100), introducing NAM leads to the most significant improvement, confirming that dynamically adjusting neurons enhances learning under class imbalance. GR further improves accuracy, particularly in CIFAR-100, showing its effectiveness in balancing gradient updates for minority classes. WS stabilizes training after neuron modifications, improving performance slightly. Finally, GDA provides additional gains, ensuring diversity in new neurons and preventing redundancy.

The full model achieves the best results (75.4\% on CIFAR-10, 41.8\% on CIFAR-100), with a total improvement of 3.2\% and 3.9 over the baseline. These results confirm that each component performs a complementary role, and their combination leads to optimal performance in class-imbalanced learning.

\section{Conclusion}
\label{sec:Conc}

In this paper, we proposed a novel approach to address class imbalance by dynamically adjusting the number of neurons during training. Inspired by neuroplasticity in the human brain, our method selectively increases neurons that contribute to the representation of minority classes while pruning those that are less significant. 
This adaptive strategy enables more effective learning of underrepresented classes while maintaining the overall stability of the model.
Experimental evaluations on three benchmark datasets (Imbalanced CIFAR-10/100, ImageNet-LT, and iNaturalist-2018) using five different architectures demonstrated that our approach consistently outperforms conventional fixed-structure networks. Notably, our method enhances minority-class accuracy without degrading overall performance and can be effectively integrated with existing imbalance-handling techniques. Furthermore, its applicability to both convolutional and Transformer-based architectures underscores its versatility.
Future work includes optimizing the scheduling of neuron modifications, improving computational efficiency, and extending the approach to other domains with severe class imbalance, such as medical imaging and anomaly detection. 
{
    \small
    \bibliographystyle{ieeenat_fullname}

}


\end{document}